\documentclass[runningheads]{llncs}
\usepackage{tabularx}
\usepackage{float}
\usepackage{booktabs} 
\usepackage{multirow} 
\usepackage{colortbl}   
\usepackage{xcolor}     
\usepackage{graphicx}   
\usepackage{color}
\usepackage{amsmath}
\usepackage{amssymb}
\usepackage{cleveref}
\usepackage[T1]{fontenc}
\usepackage{subcaption}
\usepackage{graphicx,verbatim}
\usepackage{color}
\usepackage{bm}

\begin{document}
\title{The Texture-Shape Dilemma: Boundary-Safe Synthetic Generation for 3D Medical Transformers}

\author{Jiaqi Tang\inst{1}\thanks{These authors contributed equally to this work.} \and
Weixuan Xu\inst{2}\protect\footnotemark[1] \and
Shu Zhang\inst{3} \and
Fandong Zhang\inst{3} \and
Qingchao Chen\inst{1}\thanks{Corresponding author.}}

\authorrunning{J. Tang et al.}

\institute{Peking University \\
\email{jiaqi\_tang@hsc.pku.edu.cn, qingchao.chen@pku.edu.cn} \and
Central South University \\
\email{8208231610@csu.edu.cn} \and
Deepwise Ltd. \\
\email{\{zhangshu, zhangfandong\}@deepwise.com}}

  
\maketitle              
\begin{abstract}
Vision Transformers (ViTs) have revolutionized medical image analysis, yet their data-hungry nature clashes with the scarcity and privacy constraints of clinical archives. Formula-Driven Supervised Learning (FDSL) has emerged as a promising solution to this bottleneck, synthesizing infinite annotated samples from mathematical formulas without utilizing real patient data. However, existing FDSL paradigms rely on simple geometric shapes with homogeneous intensities, creating a substantial gap by neglecting tissue textures and noise patterns inherent in modalities like CT and MRI. In this paper, we identify a critical optimization conflict termed boundary aliasing: when high-frequency synthetic textures are naively added, they corrupt the image gradient signals necessary for learning structural boundaries, causing the model to fail in delineating real anatomical margins. To bridge this gap, we propose a novel Physics-inspired Spatially-Decoupled Synthesis framework. Our approach orthogonalizes the synthesis process: it first constructs a gradient-shielded buffer zone based on boundary distance to ensure stable shape learning, and subsequently injects physics-driven spectral textures into the object core. This design effectively reconciles robust shape representation learning with invariance to acquisition noise. Extensive experiments on the BTCV and MSD datasets demonstrate that our method significantly outperforms previous FDSL, as well as SSL methods trained on real-world medical datasets, by 1.43\% on BTCV and up to 1.51\% on MSD task, offering a scalable, annotation-free foundation for medical ViTs. The code will be made publicly available upon acceptance.


\end{abstract}

\section{Introduction}

Vision Transformers (ViTs)~\cite{liu2021swin} have transformed medical image analysis but suffer from a lack of inductive biases, making them notoriously data-hungry and prone to overfitting\cite{dosovitskiy2020image}.
While Self-Supervised Learning (SSL) leverages unlabeled data to mitigate this~\cite{chen2023masked,Wu2024Voco,Tang2022SwinUNETR}, it remains tethered to clinical archives, perpetuating privacy constraints and dataset-specific biases. Consequently, there is a critical need for pre-training paradigms that initialize ViTs effectively without relying on the restrictive pipeline of real-world data collection.

In response to these limitations, Formula-Driven Supervised Learning (FDSL) has emerged as a compelling, privacy-preserving alternative. Instead of relying on natural data, FDSL synthesizes massive pairs of images and dense annotation masks directly from mathematical formulas, enabling models to be pre-trained in a fully supervised manner.~\cite{shinoda2023segrcdb,takashima2023visual,nakamura2023pre,kataoka2021formula}. 
By learning geometric primitives from synthetic patterns, FDSL offers a theoretically almost infinite source of auto-annotated diversity that completely circumvents the bottlenecks of human annotation and patient privacy. 
Previous works in natural images FDSL, such as TileDB~\cite{kataoka2021formula} and RCDB~\cite{kataoka2022replacing}, have demonstrated that pre-training on synthetic contours alone can rival performance achieved with massive real-world datasets like ImageNet-21k~\cite{deng2009imagenet}. 
Recently, this concept has been extended to the medical domain by PrimGeoSeg~\cite{tadokoro2023pre}, which utilizes 3D geometric primitives (e.g. cylinders, cones) to instill spatial awareness in volumetric networks.

\begin{figure}[t]
    \centering
    \includegraphics[width=0.99\linewidth, height=0.5\linewidth]{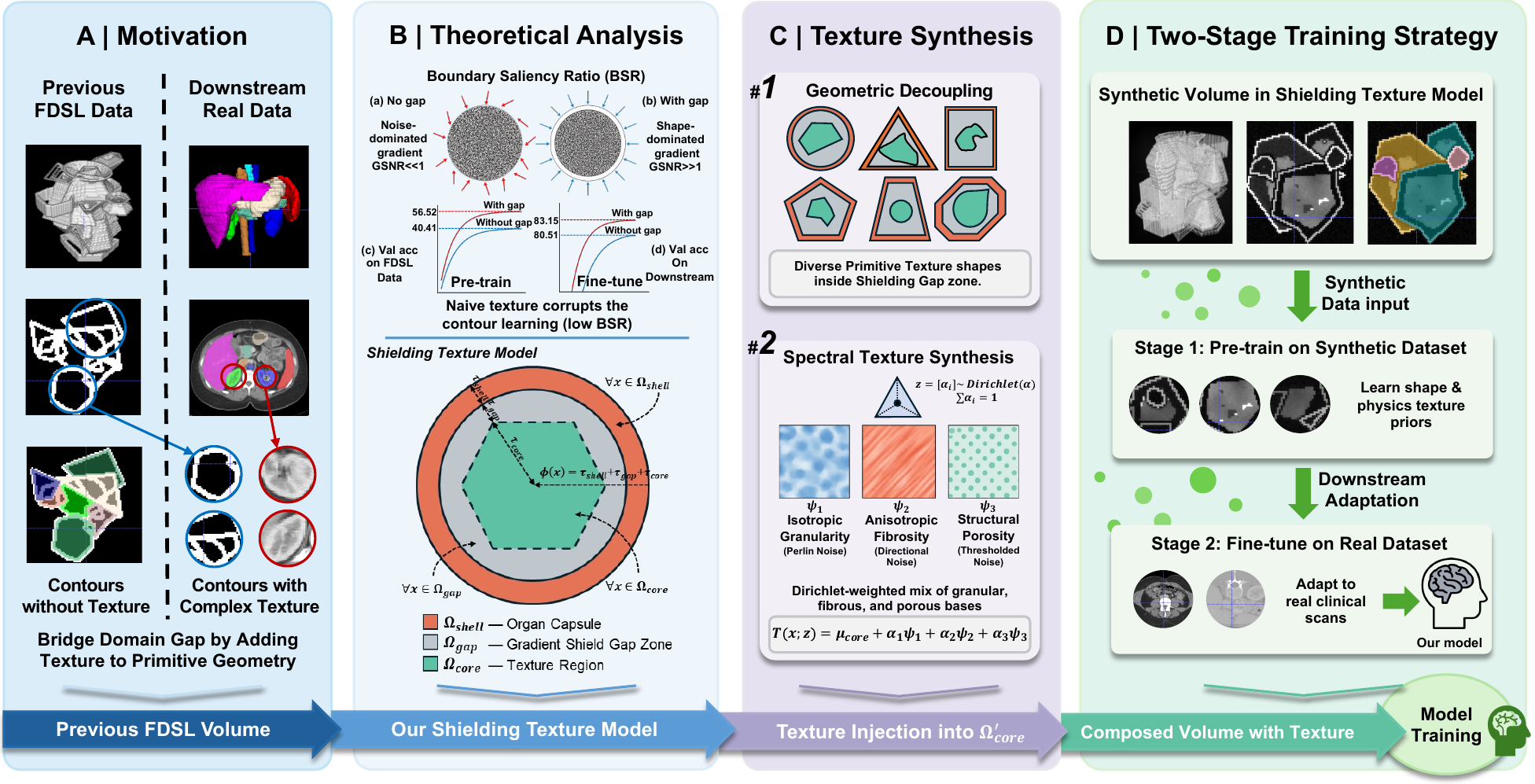}
    \caption{Overview of the proposed physics-inspired spatially-decoupled synthesis framework. From left to right: (A) Existing FDSL relies on homogeneous geometric primitives, resulting in a texture gap to real clinical data. (B) Naive texture injection leads to boundary aliasing, where stochastic gradients corrupt structural cues; a gradient-shielded buffer is constructed to preserve boundary signals. (C) Physics-driven textures are injected into a geometrically decoupled core via Dirichlet-weighted spectral mixing. (D) The composed volumes enable synthetic pre-training followed by downstream fine-tuning on real datasets.}
    \label{fig:framework}
\end{figure}

Despite these advances, a gap persists between current FDSL paradigms and the intrinsic characteristics of medical imaging. 
As shown in Figure~\ref{fig:framework}.A, Existing methods, including PrimGeoSeg~\cite{tadokoro2023pre}, rely predominantly on geometric contours with \textit{uniform, constant-intensity regions}, completely devoid of internal texture. While this representation suffices for learning structural boundaries, it fails to consider the heterogeneous tissue patterns and acquisition noise inherent in modalities like CT and MRI. 
However, bridging this gap is non-trivial. As shown in Figure~\ref{fig:framework}.B, in the prior experiment, we observe that naively adding high-frequency textures can corrupt the image gradient signals essential for boundary learning, leads to substantial degradation in FDSL pre-training performance (dropping from 56\% to 40\%), a phenomenon we formally defined as \textit{boundary aliasing} in our theoretical analysis (Section~\ref{subsec:gradient_analysis}).
Consequently, models pre-trained on such conflicting signals struggle to adapt to the nuanced intensity distributions of real pathology, limiting the transferability of FDSL to downstream tasks (downstream performance drops from 83\% to 80\%).

To address these challenges, we propose a novel \textbf{physics-inspired spatially-decoupled synthesis framework} designed to bridge the domain gap between mathematical formulas and realistic medical acquisition. Our core insight is to orthogonalize the synthesis process into two complementary modules: the \textbf{Shielding Texture Model} and \textbf{Spatially-Decoupled Texture Synthesis}.
Unlike prior works that simply overlay noise, we first introduce a \textit{Shielding Texture Model} that constructs a gradient-shielded buffer zone via distance transforms. This topological constraint ensures that the gradient signals essential for boundary learning remain pristine and unaffected by internal high-frequency artifacts.
Subsequently, we implement \textit{Geometric Decoupling} combined with \textit{Spectral Texture Synthesis}. By injecting physics-driven textures into a geometrically independent core, we mimic the biophysical appearance of real anatomy while mathematically guaranteeing that texture patterns do not interfere with boundary optimization.

We validate our approach on the BTCV~\cite{landman2015miccai} and MSD~\cite{antonelli2022medical} datasets using standard UNETR~\cite{hatamizadeh2022unetr} and SwinUNETR~\cite{hatamizadeh2022swinunetr} architectures. 
Extensive experiments demonstrate that our method significantly outperforms both training from scratch and state-of-the-art FDSL baselines by 1.43\% on BTCV and up to 1.51\% on challenging MSD tasks. 
Our results suggest that incorporating physics-based texture into synthetic pre-training effectively reduces the reliance on large-scale annotated medical datasets, offering a scalable and privacy-preserving path for training medical ViTs.

\section{Methodology}
\label{sec:method}

Previous FDSL methods, such as PrimGeoSeg~\cite{tadokoro2023pre}, rely exclusively on boundary cues and fail to capture the complex textural interplay inherent in real medical scans. 
Motivated by our theoretical analysis in Sec.~\ref{subsec:gradient_analysis}, which proposes the Boundary Saliency Ratio (BSR) to quantify how stochastic texture gradients corrupt boundary localization signals, we propose a \textit{Physics-Inspired Spatially-Decoupled Synthesis Framework}. 
As illustrated in Fig.~\ref{fig:framework}, our framework consists of two novel components corresponding to the identified challenges: 
\textbf{(i) Shielding Texture Model} (Sec.~\ref{subsec:shileding_model}), which constructs a gradient-shielded buffer zone to ensure boundary robustness and preserve high BSR; and 
\textbf{(ii) Spatially-Decoupled Texture Synthesis} (Sec.~\ref{subsec:texture_gen}), which employs \textit{Geometric Decoupling} and \textit{Physics-Driven Spectral Mixing} that mimic tissue-specific appearance on the object core. 
This design effectively reconciles textural heterogeneity with geometric optimization stability, enabling the network to learn robust shape priors without interference from high-frequency noise artifacts while bridging the texture gap.

\subsection{Boundary Aliasing: The Texture-Boundary Conflict}
\label{subsec:gradient_analysis}
We analyze how stochastic texture fields affect the \textit{spatial intensity gradients} of the input images, which serve as the fundamental cues for neural networks to detect anatomical boundaries. This analysis acts as a diagnostic to expose why naive texture synthesis leads to feature corruption.

\noindent\textbf{Setup.} Let $\Omega \subset \mathbb{R}^3$ denote the 3D image domain. A geometric primitive is represented by a binary occupancy map $M : \Omega \to \{0, 1\}$. We analyze the spatial derivative $\nabla_{\mathbf{x}}$ at the boundary $\partial M$. Working in discrete voxel coordinates (unit length $\Delta x = 1$), an intensity contrast $\Delta \mu$ across a boundary implies a spatial gradient magnitude of $\Delta \mu$. Following previous FDSL methods~\cite{tadokoro2023pre}, we assume a uniform background $B(\mathbf{x}) \equiv \mu_{bg}$, yielding $\nabla_{\mathbf{x}} B = \mathbf{0}$.

For a clean geometric input $\mathbf{X}_{geo}$, the spatial gradient at a boundary voxel $\mathbf{x}_b$ provides a pristine signal for edge detection, well-approximated by the surface normal:
\begin{equation}
    \nabla_{\mathbf{x}} \mathbf{X}_{geo}(\mathbf{x}_b) \approx \Delta \mu \, \hat{\mathbf{n}}(\mathbf{x}_b) + \mathcal{O}(\delta),
    \label{eq:clean_gradient}
\end{equation}
where $\hat{\mathbf{n}}$ is the outward unit normal. This vector $\nabla_{\mathbf{x}} \mathbf{X}_{geo}$ represents the ideal \textit{geometric cue} that the network should learn to encode shape priors.

\noindent\textbf{Signal Corruption under Naive Texture Synthesis.} A naive approach overlays stochastic textures $T(\mathbf{x})$ directly onto the shape: $\mathbf{X}_{naive}(\mathbf{x}) = M(\mathbf{x}) T(\mathbf{x}) + (1 - M(\mathbf{x})) B(\mathbf{x})$. Expanding the spatial gradient via the discrete product rule reveals a conflict:
\begin{equation}
    \nabla_{\mathbf{x}} \mathbf{X}_{naive}(\mathbf{x}_b) \approx \underbrace{[T(\mathbf{x}_b) - B(\mathbf{x}_b)] \, \hat{\mathbf{n}}(\mathbf{x}_b)}_{\text{Geometric Signal}} + \underbrace{M(\mathbf{x}_b) \nabla_{\mathbf{x}} T(\mathbf{x}_b)}_{\text{Texture Interference}}.
    \label{eq:gradient_decomp}
\end{equation}
Since medical textures often exhibit high-frequency variations (e.g., trabecular bone), the interference term $\|\nabla_{\mathbf{x}} T\|$ is frequently comparable in magnitude to the boundary contrast, making the texture gradients overwhelm the geometric normal vector and confuse the network's feature extractors.
To quantify this conflict, we define the \textit{Boundary Saliency Ratio (BSR)} at boundary voxels:
\begin{equation}
    \text{BSR}(\mathbf{x}_b) = \frac{\left| T(\mathbf{x}_b) - B(\mathbf{x}_b) \right|^2}{\mathbb{E}_{\mathbf{z}} [\|\nabla_{\mathbf{x}} T(\mathbf{x}_b; \mathbf{z})\|^2] + \epsilon}.
    \label{eq:bsr}
\end{equation}
When $\text{BSR}(\mathbf{x}_b) \ll 1$, the input features are dominated by stochastic texture artifacts rather than structural boundary, a condition we term \textit{boundary aliasing}. This implies that to learn robust shape priors, we must enforce $\|\nabla_{\mathbf{x}} T(\mathbf{x})\| \to 0$ within a local neighborhood of $\partial M$. This theoretical insight directly motivates our \textit{Shielding Texture Model}, which constructs a gradient-shielded buffer zone to guarantee high BSR while preserving internal textural diversity.

\subsection{Shielding Texture Model}
\label{subsec:shileding_model}

To resolve the conflict identified above, we introduce a spatial decoupling mechanism. Instead of global texturing, we orthogonalize the volume into functional strata based on the Euclidean Distance Transform (EDT), $\mathcal{D}_M(\mathbf{x}) = \min_{\mathbf{y} \in \partial M} \|\mathbf{x} - \mathbf{y}\|_2$.
We partition the foreground $\Omega_{obj}$ into three distinct regions:
\begin{equation}
    \Omega_{k} = 
    \begin{cases} 
        \Omega_{shell} & \text{if } 0 < \mathcal{D}_M(\mathbf{x}) \le \tau_{shell} \\
        \Omega_{gap}   & \text{if } \tau_{shell} < \mathcal{D}_M(\mathbf{x}) \le \tau_{shell} + \tau_{gap} \\
        \Omega_{core}  & \text{if } \mathcal{D}_M(\mathbf{x}) > \tau_{shell} + \tau_{gap}
    \end{cases},
\end{equation}
where $\Omega_{shell}$ mimics the organ capsule with constant intensity $\mu_{shell}$.

In this mode, the region $\Omega_{gap}$ acts as a critical \textit{gradient shield}. By enforcing a constant intensity transition $\mu_{gap}$ within this buffer, we strictly impose $\nabla \mathbf{X}(\mathbf{x}) = \mathbf{0}, \forall \mathbf{x} \in \Omega_{gap}$. 
Crucially, the buffer width $\tau_{gap}$ is set to exceed the kernel size of the network's first-layer gradient operators (e.g., $\tau_{gap} \ge 2$ for $3\times3$ kernels). This topological constraint theoretically ensures $\text{BSR}(\mathbf{x}_b) \to \infty$, allowing the network to learn shape priors driven purely by deterministic contrast without stochastic corruption from the texture core.

\subsection{Spatially-decoupled Physics-inspired Texture Synthesis}
\label{subsec:texture_gen}

While the spatial decoupling guarantees boundary stability, simple randomized textures inside $\Omega_{core}$ can lead to trivial solutions such as overfitting to concentric patterns. We further introduce a physics-driven modeling approach that enhances realism while preventing such shortcuts through \textit{Geometric Decoupling} and \textit{Spectral Texture Synthesis}.

\noindent\textbf{Geometric Decoupling.}
A naive erosion of the outer shape $M$ to define the texture region creates a strong spatial correlation that networks can exploit. To break this dependency, we generate the texture-bearing region $\Omega'_{core}$ using an independent geometric primitive $M_{inner}$ (e.g., a prism within a cylinder). The final core region is defined as the intersection of this transformed primitive and the EDT-safe zone:
\begin{equation}
    \Omega'_{core} = \mathcal{A}(M_{inner}) \cap \{\mathbf{x} : \phi(\mathbf{x}) > \tau_{shell} + \tau_{gap}\},
\end{equation}
where $\mathcal{A}(\cdot)$ denotes a random affine transformation. This ensures that the internal texture boundary $\partial \Omega'_{core}$ is spatially decorrelated from the organ boundary $\partial M$, forcing the network to learn global shape semantics rather than local intensity transitions.

\noindent\textbf{Spectral Texture Synthesis.}
To bridge the texture gap with medical scans, we model the texture field $\mathcal{T}(\mathbf{x})$ not as simple Gaussian noise, but as a convex combination of biophysical archetypes. We introduce a spectral mixing function:
\begin{equation}
    \mathcal{T}(\mathbf{x}; \mathbf{z}) = \mu_{core} + \sum_{i=1}^{3} \alpha_i \Psi_i(\mathbf{x}), \quad \text{s.t. } \sum \alpha_i = 1,
\end{equation}
where $\mathbf{z}=\{\alpha_i\}$ are mixing weights drawn from a Dirichlet distribution. The basis functions $\Psi_i$ capture distinct tissue properties: (1) \textit{Isotropic Granularity} via multi-scale Perlin noise for parenchyma; (2) \textit{Anisotropic Fibrosity} via directionally scaled noise fields $\Psi(\mathbf{S}_{\mathbf{v}}\mathbf{x})$; and (3) \textit{Structural Porosity} via thresholded noise fields to mimic trabecular bone.
By confining this complex, high-frequency signal strictly within $\Omega'_{core}$, we achieve the dual goal of synthesizing texturally realistic volumes while maintaining the zero-gradient guarantee at the boundary critical for geometric learning.

\begin{table}[tbp]
\centering
\caption{Comparison of segmentation performance on the BTCV dataset.}
\label{tab:btcv_results}

\resizebox{\textwidth}{!}{%
\begin{tabular}{l|c|c|c|ccccccccccccc}
\toprule
\textbf{Pre-training} & \textbf{PT Num} & \textbf{Type} & \textbf{Avg.} & Spl & RKid & LKid & Gall & Eso & Liv & Sto & Aor & IVC & Veins & Pan & rad & lad \\
\midrule

\multicolumn{17}{l}{\textit{UNETR}} \\

Scratch & 0 & -- & 75.86 
& 93.02 & 89.79 & 90.60 & 51.34 & 71.12 
& 95.02 & 72.90 & 82.95 & 76.71 
& 69.05 & 76.29 & 62.02 & 55.33 \\

PrimGeoSeg & 5K & FDSL & 78.90 
& 93.19 & 93.47 & 93.35 & 54.49 & 68.91 
& 95.88 & 78.83 & 87.82 & 82.47 
& 72.73 & 80.08 & 65.06 & 59.35 \\
\rowcolor{lightgray}
Ours & 5K & FDSL & \textbf{79.61} & 93.47 & 93.10 & 92.86 & 60.50 & 69.08 & 95.82 & 80.09 & 90.24 & 83.42 & 71.49 & 79.39 & 63.93 & 61.49 \\

\midrule

\multicolumn{17}{l}{\textit{SwinUNETR}} \\
Scratch & 0 & -- & 80.12 & 95.32 & 94.01 & 93.51 & 61.17 & 73.83 & 94.40 & 71.19 & 88.13 & 85.20 & 73.12 & 80.60 & 67.15 & 63.88 \\
PrimGeoSeg & 5K & FDSL & 80.08 & 95.25 & 93.76 & 93.79 & 61.23 & 74.47 & 96.04 & 80.64 & 89.39 & 83.91 & 74.27 & 77.83 & 64.13 & 56.36 \\

\rowcolor{lightgray}
Ours & 5K & FDSL & \textbf{81.51} & 95.72 & 94.11 & 93.94 & 56.72 & 74.14 & 96.72 & 84.16 & 90.66 & 85.32 & 74.20 & 82.40 & 66.23 & 65.28 \\

\end{tabular}%
}
\end{table}

\begin{table}[htbp]
\centering
\caption{Comparison of segmentation performance on several MSD datasets.}
\label{tab:msd_results}
\resizebox{\textwidth}{!}{
\begin{tabular}{l cc cc cc }
\toprule
\multirow{3}{*}{\textbf{Method}} 
& \multicolumn{6}{c}{\textbf{MSD Dataset}} \\
\cmidrule{2-7}
& \multicolumn{2}{c}{\textbf{Task 02}} 
& \multicolumn{2}{c}{\textbf{Task 06}} 
& \multicolumn{2}{c}{\textbf{Task 09}}  \\
\cmidrule{2-3} \cmidrule{4-5} \cmidrule{6-7}
& \textbf{UNETR} & \textbf{SwinUNETR} 
& \textbf{UNETR} & \textbf{SwinUNETR} 
& \textbf{UNETR} & \textbf{SwinUNETR} \\
\midrule
Scratch       
& 95.14 & 95.34 
& 70.68 & 76.14 
& 92.52 & 95.92  \\

PrimGeoSeg    
& 95.61 & 95.77 
& 78.81 & 80.39 
& 96.30 & 96.96  \\

\rowcolor{lightgray}
\textbf{Ours}   
& \textbf{95.91} & \textbf{96.29}  
& \textbf{79.20} & \textbf{81.47}  
& \textbf{96.63} & \textbf{98.47}  \\

\bottomrule
\end{tabular}
}
\end{table}

\section{Experiments and Results}

\subsection{Experimental Setup}

\noindent
\textbf{Tasks and datasets:}
In this part, we conduct experiments on the BTCV~\cite{landman2015miccai} and MSD~\cite{antonelli2022medical} datasets to evaluate the proposed method. The BTCV dataset contains 30 CT volumes with multiple organ annotations.
For MSD, we evaluate on Task02 (Heart), Task06 (Lung) and Task09 (Spleen). Following previous methods~\cite{tadokoro2023pre}, these datasets are divided into training and validation subsets using the 80/20 split, and all experiments are conducted in an offline setting.

\noindent
\textbf{Implementation details:}
During pre-training, we use a ROI size of $96 \times96 \times 96$, a batch size of 4, a learning rate of 0.0002, and a weight decay of 0.00001. The model is optimized using AdamW with a warmup cosine learning rate scheduler with training iterations set to $25000$. 
For fine-tuning, we follow the standard hyperparameter settings used in prior works~\cite{tadokoro2023pre} with a batch size of 4. Specifically, the SwinUNETR~\cite{hatamizadeh2022swinunetr} settings on BTCV follow the configurations reported in  \cite{tang2022selfsupervised}, while the hyperparameter settings for MSD follow the established protocol described in \cite{hatamizadeh2022unetr}.

\subsection{Results}
\label{sec:analysis}

\textbf{Performance on Full Datasets}
As shown in Table~\ref{tab:btcv_results} and~\ref{tab:msd_results}, our method consistently outperforms training from scratch and the geometry-only pre-training baseline on BTCV and MSD tasks under both UNETR and SwinUNETR. On BTCV, the average Dice with SwinUNETR improves by $1.39$ over Scratch and $1.43$ over the baseline. Improvements are observed across most organs, including thin structures such as pancreas and adrenal glands. On MSD, the gain is more evident on Task06, which contains stronger appearance variation and boundary ambiguity, achieving an improvement of $5.33$ over Scratch and $1.08$ over the PrimGeoSeg baseline. 

\noindent
\textbf{Comparison with Other Self-Supervised Learning Methods}
We compare our synthetic pre-training strategy with representative SSL~\cite{wang2023swinmm,tang2022selfsupervised} methods pre-trained on real CT datasets on RadGenome Dataset~\cite{zhang2024radgenome} on BTCV dataset. While SwinUNETR SSL shows slight improvements, our method achieves 81.51 Dice, outperforming both real-data SSL methods and FDSL baselines.

\begin{table}[htp]
\centering
\caption{Comparison with SSL methods pre-trained on RadGenome Dataset~\cite{zhang2024radgenome}}
\label{tab:ssl_comparison}
\begin{tabular}{l|c|c|c}
\toprule
\textbf{Method} & \textbf{Pre-train Data} & \textbf{Data Amount} & \textbf{Avg. Dice} \\
\midrule
SwinMM~\cite{wang2023swinmm} & Real (CT) & 5,000 & 76.72 \\
SwinUNETR~\cite{tang2022selfsupervised} & Real (CT) & 5,000 & 80.56 \\
\midrule
\rowcolor{gray!20} \textbf{Ours} & \textbf{Synthetic} & \textbf{5,000} & \textbf{81.51} \\
\bottomrule
\end{tabular}
\end{table}

\subsection{Analysis}
We further analyze four key factors in our method: dataset scale, texture source, boundary gap width and backbone architecture. For the first three factors, experiments are conducted on BTCV with SwinUNETR. Results are summarized in Table~\ref{tab:ablation-all}. For backbone analysis, we further compare UNet with transformer-based architectures.

\noindent
\textbf{Scaling Effects}
We vary the number of synthetic volumes. Increasing the scale from 500 to 5,000 improves Dice from 80.16 to 81.51. Further scaling to 15,000 and 50,000 leads to 82.26 and 82.40 respectively. The result improves steadily with larger synthetic datasets and the gain becomes smaller at larger scales.

\begin{table*}[tp]
\centering
\caption{Analysis on the BTCV dataset with the SwinUNETR under different settings, including pre-training dataset scales, texture types, and gap widths.}
\label{tab:ablation-all}

\begin{tabular*}{\textwidth}{@{\extracolsep{\fill}}lc lc lc}
\toprule
\multicolumn{2}{c}{\textbf{Dataset Scale}} 
& \multicolumn{2}{c}{\textbf{Texture Type}} 
& \multicolumn{2}{c}{\textbf{Gap Width}} \\

\cmidrule(lr){1-2}\cmidrule(lr){3-4}\cmidrule(lr){5-6}

Scale & Avg. Dice 
& Type & Avg. Dice
& Width & Avg. Dice \\

\midrule
500   & 80.16 
& fruit & 80.39 
& $w=0$ & 80.02 \\

5000  & 81.51 
& trabecular  & 80.56
& $w=1$ & 80.22 \\

15000 & 82.26 
& fibrous & 81.10 
& $w=5$ & 80.55 \\

50000 & \textbf{82.40 }
& granular & 80.41 
& $w=9$ & \textbf{80.59} \\

--    & --    
& \textbf{ours} & \textbf{81.51} 
& $w=13$ & 80.15 \\

\bottomrule
\end{tabular*}

\end{table*}

\noindent
\textbf{The Texture Selection Analysis}
In this part, we evaluate the impact of texture selection. In addition to the single-type physical simulation, we introduce real fruit textures~\cite{kuLeuvenXray} for comparison. Different texture sources lead to noticeable performance variations. Compared with single-type and fruit-based textures, our texture design achieves the highest Dice score, suggesting that effective pre-training depends more on structural consistency than appearance diversity.

\noindent
\textbf{The Gap Width Analysis}
We vary the black gap width $w$ between boundary regions and inner textures. Removing the gap results in 80.02 Dice. Increasing the gap width consistently improves performance. The best result is obtained at $w=9$ with 80.59, indicating that explicitly separating boundary regions encourages the model to learn more discriminative structure-aware representations.

\noindent
\textbf{Backbone Generalization Analysis}
In addition to the experiments reported in Table~\ref{tab:btcv_results}, we further evaluate our method on a standard 3D U-Net architecture. Training from scratch achieves \textbf{78.87} Dice, while the baseline improves the performance to \textbf{80.64}. Our method further boosts the Dice score to \textbf{81.94}. These results demonstrate that the proposed synthetic pre-training strategy is not limited to transformer-based architectures, but also effectively enhances convolutional backbones.

\section{Conclusion}
In this paper, we bridge the domain gap in FDSL via a novel physics-inspired spatially-decoupled synthesis framework. By identifying and resolving the boundary aliasing conflict, our approach orthogonalizes the learning of robust geometric priors from heterogeneous textural modeling. Extensive experiments demonstrate that this shielding-and-decoupling paradigm enables our model not only to surpass state-of-the-art FDSL and SSL baselines but, crucially, to achieve performance parity with pre-training on large-scale real medical datasets. Consequently, our method offers a scalable, privacy-preserving foundation for initializing medical Vision Transformers without the constraints of clinical data collection.

\bibliographystyle{splncs04}
\bibliography{cite}



\end{document}